\documentclass[letterpaper, 10 pt, conference]{ieeeconf}  

\IEEEoverridecommandlockouts                                                                                        
\usepackage{graphics} \usepackage{epsfig} \usepackage{amsmath} \usepackage{amssymb}  
\usepackage{balance}
\usepackage{color}
\usepackage{comment}
\usepackage{caption}
\usepackage{multirow}
\usepackage{hyperref}
\usepackage{siunitx}
\usepackage{dblfloatfix}    
\let\labelindent\relax
\usepackage{enumitem}

\newlist{propertylist}{enumerate}{1}
\setlist[propertylist]{label=P\arabic*:}

\usepackage[labelformat=simple]{subcaption}

\DeclareCaptionLabelSeparator{periodspace}{.\quad}
\newcommand{\figref}[1]{Fig.~\ref{fig:#1}}

\newcommand{\vy}{\mathbf{y}}

\newcommand{\R}{\mathbb{R}}

\newcommand{\cE}{\mathcal{E}}
\newcommand{\cF}{\mathcal{F}}
\newcommand{\cO}{\mathcal{O}}
\newcommand{\cP}{\mathcal{P}}
\newcommand{\cS}{\mathcal{S}}
\newcommand{\cW}{\mathcal{W}}
\renewcommand{\ij}{^{(i,j)}}
\newcommand{\ji}{^{(j,i)}}
\renewcommand{\th}{^{\text{th}}}

\newcommand{\bmat}[1]{\begin{bmatrix} #1 \end{bmatrix}}

\DeclareMathOperator*{\conv}{\mathbf{conv}}
\DeclareMathOperator*{\cost}{\text{cost}}
\DeclareMathOperator*{\diag}{\mathbf{diag}}

\graphicspath{{images/}{../cs646/}{../../figures/}}

\title{\LARGE \bf
Downwash-Aware Trajectory Planning for Large Quadrotor Teams
}
\author{James A. Preiss, Wolfgang H\"onig, Nora Ayanian, and Gaurav S. Sukhatme\thanks{All authors are with the Department of Computer Science, University of Southern California, Los Angeles, CA, USA.}\thanks{Email: {\tt\footnotesize \{japreiss, whoenig, ayanian, gaurav\}@usc.edu}}\thanks{This work was partially supported by the ONR grants N00014-16-1-2907 and N00014-14-1-0734, and the ARL grant W911NF-14-D-0005.}}
\begin{document}
\maketitle
\thispagestyle{empty}
\pagestyle{empty}

\begin{abstract}
We describe a method for formation-change trajectory planning for large quadrotor teams in obstacle-rich environments. Our method decomposes the planning problem into two stages: a discrete planner operating on a graph representation of the workspace, and a continuous refinement that converts the non-smooth graph plan into a set of $C^k$-continuous trajectories, locally optimizing an integral-squared-derivative cost. We account for the downwash effect, allowing safe flight in dense formations. We demonstrate the computational efficiency in simulation with up to 200 robots and the physical plausibility with an experiment with 32 nano-quadrotors.
Our approach can compute safe and smooth trajectories for hundreds of quadrotors in dense environments with obstacles in a few minutes.
\end{abstract}

\section{INTRODUCTION}
Trajectory planning is a fundamental problem in multi-robot systems.
Given a set of robots with known initial locations and a set of goal locations,
the task is to find a one-to-one goal assignment and 
a set of continuous functions that move each robot
from its start position to its goal,
while avoiding collisions and respecting dynamic limits.
Trajectory planning is a core subproblem of various applications
including search-and-rescue, inspection, and delivery.
In this work we address the \emph{unlabeled} case; in the \emph{labeled} case the goal assignment is given.

A large body of work has addressed this problem with varied discrete and continuous formulations.
However, no existing solution simultaneously satisfies the goals of completeness,
physical plausibility, optimality in time or energy usage,
and good computational performance.
In this work, we present a method that attempts to balance these goals.

\begin{figure}
\centering
\includegraphics[width=\columnwidth]{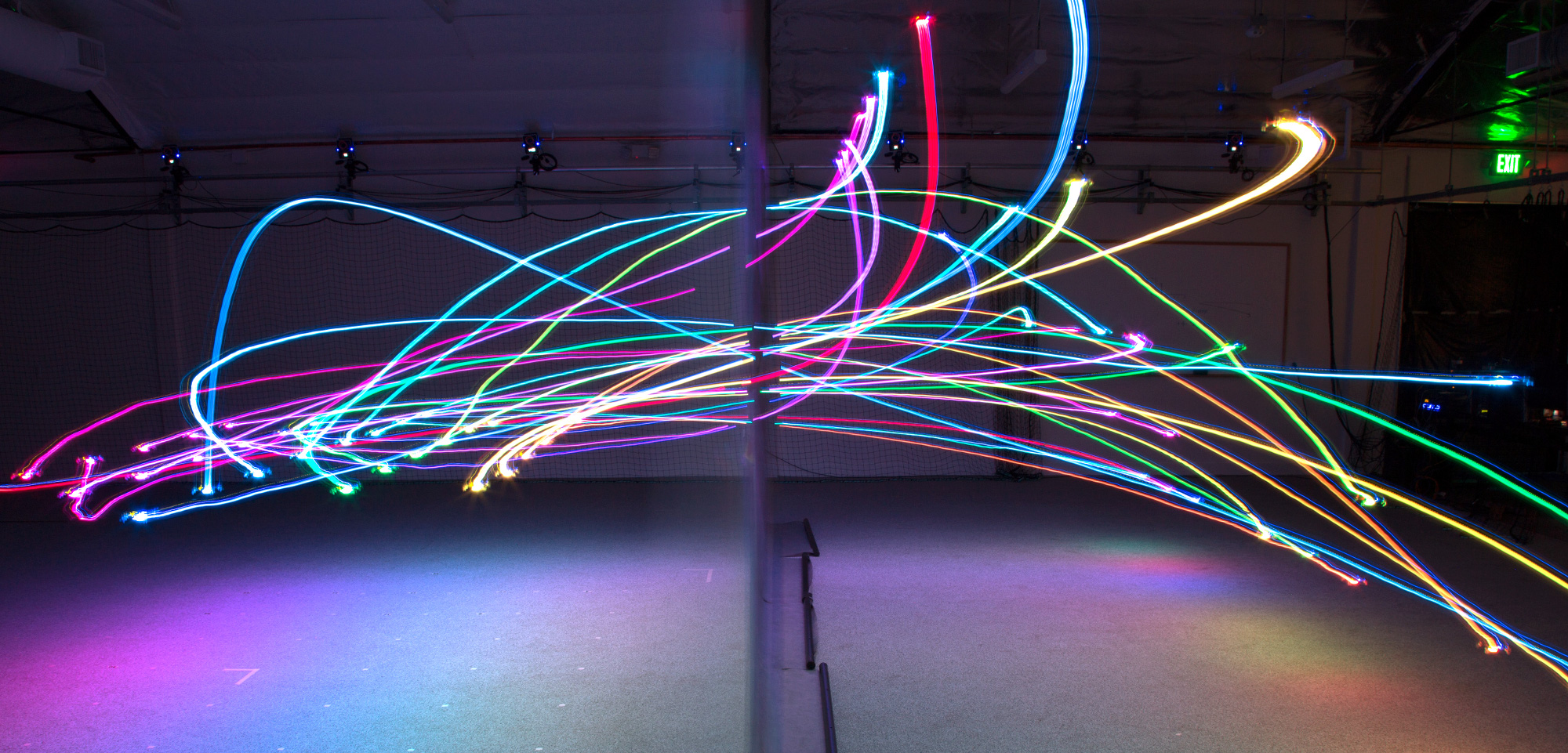}
\caption{
Long exposure of 32 Crazyflie nano-quadrotors flying through a wall with windows,
viewed from the edge of the wall.
}
\label{fig:cover}
\end{figure}

Our method uses a graph-based planner to compute a solution for a discretized version of the problem, 
and then refines this solution into smooth trajectories in a separate, decoupled optimization stage.
We directly take the downwash effect of quadrotors into account,
preserving safety during dense formation flights.
Furthermore, our method is complete with respect to the resolution of the discretization, and locally optimal
with respect to an energy-minimizing integral-squared-derivative objective function.
We also present an anytime iterative refinement scheme
that improves the trajectories within a given computational budget.
We support user-specified smoothness constraints and provide simulations with up to 200 robots and a physical experiment with 32 quadrotors, see~\figref{cover}.

\section{RELATED WORK}
A simple approach to multi-robot motion planning is to repurpose a single-robot planner
and represent the Cartesian product of the robots' configuration spaces
as a single large joint configuration space~\cite{lavalleplanning}.
Robot-robot collisions are represented as configuration-space obstacles.
However, the high-dimensional search space is computationally infeasible for large teams.

Many works have approached the problem from a graph search perspective
\cite{mstar, conflictbased}.
These methods are adept at dealing with maze-like environments and scenarios with high congestion.
Some represent the search graph implicitly~\cite{needlehaystack},
so they are not always restricted to a predefined set of points in configuration space.
However, directly interpreting a graph plan as a trajectory results in a piecewise linear path, requiring the robot to fully stop at each graph vertex to maintain dynamic feasibility.
It is possible to use these planners to resolve ordering conflicts and refine the output for execution on robots~\cite{wolfgang}.

Some authors have solved the formation change problem in a continuous setting \cite{augugliaro, mellinger-mixed},
but methods are often tightly coupled, solving one large optimization problem
in which the decision variables define all robots' trajectories.
These approaches are typically demonstrated on smaller teams
and do not scale up to the size of team in which we are interested.
Others decouple the problem but do not support the level of smoothness
in our solution \cite{chen2015decoupled}, and the authors do not show results on large teams.
The method of~\cite{DBLP:conf/rss/TurpinMMK13} is computationally 
fast, but
offsets the different trajectories in time, resulting in much longer time durations.
Velocity profile methods~\cite{velocityprofile} handle kinodynamic constraints well but are not able to fully exploit free space in the environment.
Collision-avoidance approaches \cite{rvo-lqr, orca-nonhol} let each robot plan its trajectory independently
and resolve conflicts in real time when impending collisions are detected.
These methods scale well, and their robustness against disturbances is appealing.
However, they do not provide any means to optimize the trajectories for objectives
such as time or energy use, and they 
are poorly suited
to problems in maze-like environments.

Spline-based refinement of waypoint plans was described in~\cite{tang} and~\cite{flores}.
Our method builds upon these works by adding support for three-dimensional ellipsoidal robots,
environmental obstacles, 
and an anytime refinement stage to further improve the plan after generating an initial set of smooth trajectories.
We demonstrate that our iterative refinement 
produces trajectories with significantly smoother dynamics.

\section{APPROACH}

We start by introducing the robot model, which is required to model the downwash effect. 
We then formalize the problem statement and outline our approach.
In later sections, we will discuss each part of our approach in detail.

\subsection{Robot Model}

As aerial vehicles, quadrotors have a six-dimensional configuration space.
However, as shown in \cite{mellingersnap}, quadrotors are \emph{differentially flat} in the \emph{flat outputs}
$(x, y, z, \psi)$, where $x, y, z$ is the robot's position in space and $\psi$ its yaw angle (heading).
Differential flatness implies that the control inputs needed to move the robot along a trajectory in the flat outputs
are algebraic functions of the flat outputs and a finite number of their derivatives.
Furthermore, in many applications, a quadrotor's yaw angle is unimportant and can be fixed at $\psi = 0$.
We therefore focus our efforts on planning trajectories in three-dimensional Euclidean space.

While some multi-robot planning work has considered simplified dynamics models
such as kinematic agents \cite{wolfgang} or double-integrators \cite{augugliaro},
our method produces trajectories with arbitrary smoothness up to a user-defined derivative.
This goal is motivated by \cite{mellingersnap}, where it was shown that
a continuous fourth derivative of position is necessary for physically plausible quadrotor trajectories,
because it ensures that the quadrotor will not be asked to change its motor speeds instantaneously.

\begin{figure}
\centering
\includegraphics[width=0.3\textwidth]{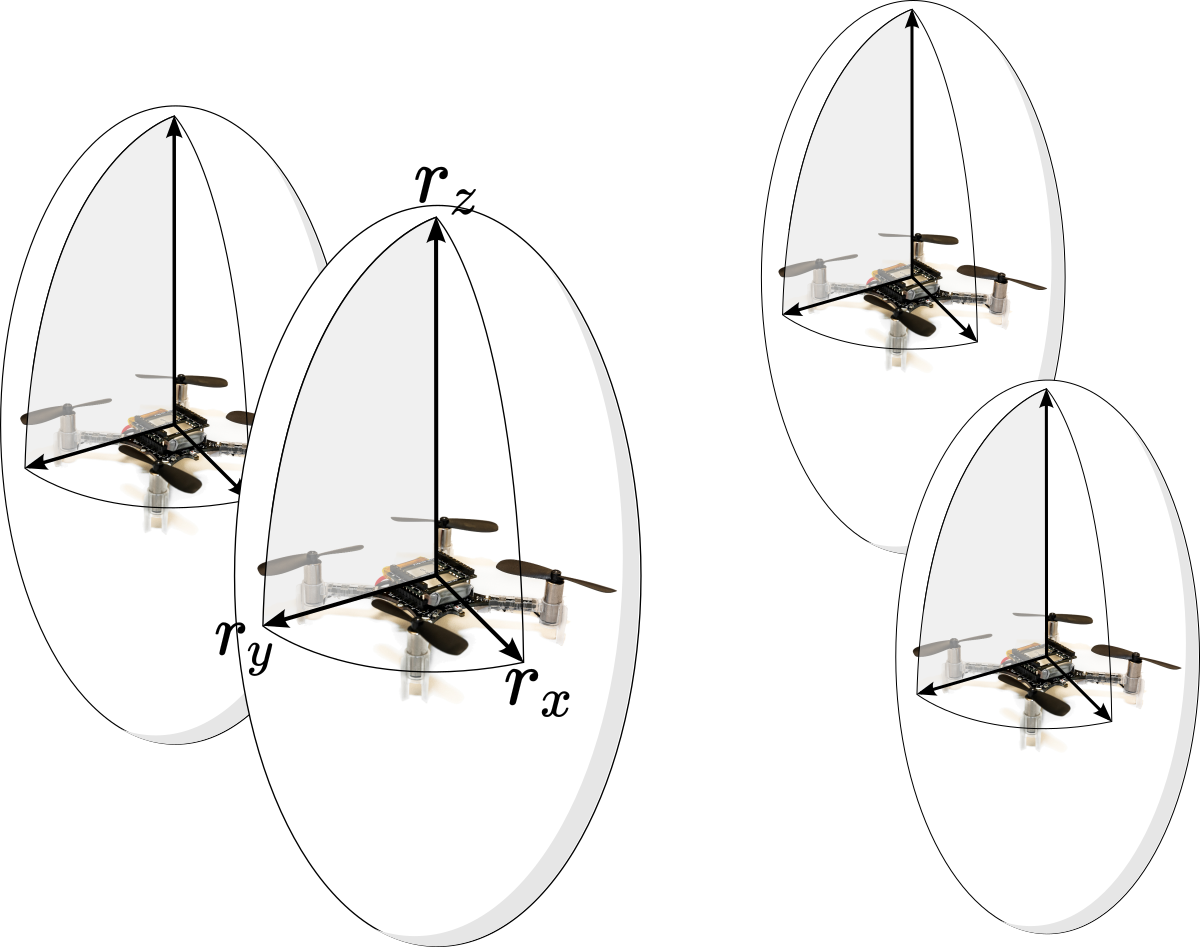}
\caption{Axis-aligned ellipsoid model of robot volume.
Tall height prevents downwash interference between quadrotors.
}
\label{fig:ellipsoid}
\end{figure}

Rotorcraft generate a large, fast-moving volume of air underneath their rotors called \emph{downwash}.
The downwash force is large enough to cause a catastrophic loss of stability when one rotorcraft flies underneath another.
We model downwash constraints by treating each robot
as an axis-aligned ellipsoid of radii \mbox{$0 < r_x = r_y \ll r_z$}, illustrated in \figref{ellipsoid}.
Empirical data collected in \cite{yeodownwash, grasp} support this model.
The set of points representing a robot at position $q \in \R^3$ is given by
\begin{equation}
\cE(q) = \{ E x + q : \|x\|_2 \leq 1\}
\end{equation}
where $E = \diag(r_x, r_y, r_z)$.
The collision-avoidance constraint between robots located at $p, q \in \R^3$ is given by
\begin{equation}
\|E^{-1}(p - q)\|_2 \geq 2.
\label{eq:collision}
\end{equation}

\subsection{Problem Statement}

\begin{figure*}[!b]
\centering
\begin{subfigure}[b]{0.2\textwidth}
\centering
\includegraphics[height=2cm]{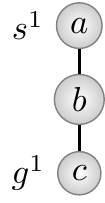}
\caption{$\mathcal{G}_E$}
\label{fig:flowgraph:env}
\end{subfigure} \hfill
\begin{subfigure}[b]{0.3\textwidth}
\centering
\includegraphics[height=1.6cm]{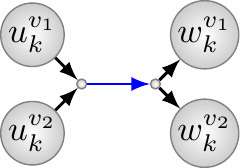}
\caption{``Gadget'' for flow-graph construction.}
\label{fig:flowgraph:gadget}
\end{subfigure}
\begin{subfigure}[b]{0.45\textwidth}
\centering
\includegraphics[height=2.5cm]{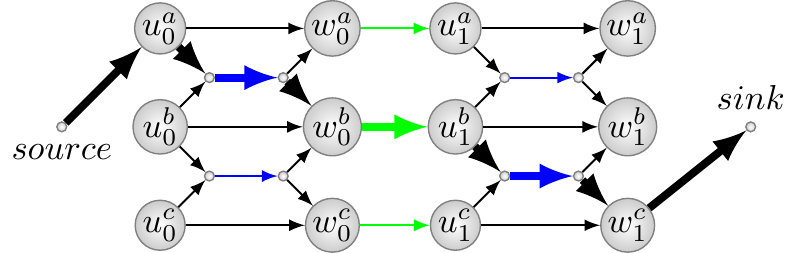}
\caption{$\mathcal{G}_F$ with $K=2$.}
\label{fig:flowgraph:example}
\end{subfigure}
\caption{Example flow-graph (Fig.~\ref{fig:flowgraph:example}) for environment shown in Fig.~\ref{fig:flowgraph:env} with a single robot.
The construction uses a graph ``gadget'' (Fig.~\ref{fig:flowgraph:gadget}) for each edge in $\mathcal{G}_E$.
The blue edges are annotated with downwash edge conflicts and the green edges are annotated with downwash vertex conflicts.
The bold arrows in Fig.~\ref{fig:flowgraph:example} show the maximum flow through the network, which can be used to compute the robots' paths.
}
\label{fig:flowgraph}
\end{figure*}

Consider a team of $N$ robots in a bounded environment containing convex obstacles $\cO_1 \dots \cO_{N_{obs}}$.
Boundaries of the environment are defined by a convex polytope $\cW$.
The free configuration space for a single robot is thus given by
\begin{equation}
\cF = \left(\cW \setminus \left(\textstyle{\bigcup_h} \cO_h\right)\right) \circleddash \cE(\mathbf{0})
\end{equation}
where $\circleddash$ denotes the Minkowski difference.

We are given a start position for each robot $s^i \in \cF$
and a set of goal positions $G \subset \cF, |G| = N$.
The start and goal inputs must satisfy the collision constraint \eqref{eq:collision} for all robot pairs.
We seek the following:

\begin{itemize}
\item
An assignment of each robot to a goal position ${g^{\phi(i)} \in G}$, where $\phi$ is a permutation of $1 \ldots N$

\item
The total time duration $T \in \R_{> 0}$
until the last robot reaches its goal

\item
For each robot $r^i$, a trajectory $f^i : [0, T] \mapsto \cF$
where $f^i(0) = s^i, \quad f^i(T) = g^{\phi(i)}$, and $f^i$ must be continuous up to a user-specified parameter $C$:
\begin{equation}\begin{split}
\frac{d^c}{dt^c} f^i(t) \text{ continuous for all } c \in \{1 \dots C\}.
\label{eq:foft}
\end{split}\end{equation}
Additionally, we require that the collision-avoidance constraint
\eqref{eq:collision} is satisfied at all times for all pairs of robots.
\end{itemize}

In the following, we present an efficient solution to the subclass of problems where all $s^i$ and $g^i$ are positions in an orthogonal grid and obstacles are cubes within that grid.

\subsection{Overview}
Our approach decomposes the formation change problem into two steps:
\emph{Discrete Planning} and \emph{Continuous Refinement}.
Discrete planning solves the goal assignment problem (generating $\phi$)
and computes a timed sequence of waypoints for each robot in a graph approximation of the environment.
Continuous refinement uses the discrete plan as a starting point
to compute a set of smooth trajectories satisfying user-supplied smoothness constraints.

We note that a major benefit of our method is its ability
to use different discrete planners.
For example, it would be possible to use a discrete planner for planning problems where the goal assignment is fixed a-priori, or where robots are split into smaller groups.

\section{DISCRETE PLANNING STAGE}

The discrete planning stage works with a grid discretization of the environment. We assume that the robots' start and goal locations are vertices of the underlying graph.

\subsection{Overview}

The discrete planning stage computes the goal assignment~$\phi$
and a path $p^i$ for each robot composed of a sequence of $K+1$ (time, position) pairs:
\begin{equation}\begin{split}
p^i = (t_0, x^i_0), (t_1, x^i_1), \dots, (t_{K}, x^i_{K})
\end{split}\end{equation}
where $0 = t_0 < t_1 < \dots < t_{K} = T$, $x^i_k \in \cF$, $x^i_0 = s^i$, and $x^i_{K} = g^{\phi(i)}$.
In between waypoints $(t_k, x^i_k)$ and $(t_{k+1}, x^i_{k+1})$,
we assume that robot $i$ travels on the line segment between $x^i_k$ and $x^i_{k+1}$,
but we do not make any assumptions about the velocity profile of the robot along that path.
We denote this line segment by $\ell^i_k$.

We require that the discrete planner supplies a plan
that satisfies the ellipsoid collision-avoidance constraint \eqref{eq:collision}
for all possible identical velocity profiles.
We also require all robots to share the same sequence of waypoint times $t_0 \dots t_K$.

In the following, we discuss one specific discrete planner
that simultaneously computes the goal assignment $\phi$
and produces waypoint sequences $p^i$ that minimize $K$.
This planner operates in a grid environment and assumes fixed timesteps, i.e. $t_{k+1} - t_k$ is equal for all $k$.
Furthermore, we require the grid size to be greater than $2r_x$.
A robot can either move to an adjacent grid cell or stay at its current location each step. 
At all timesteps, and during movements, the planner must ensure that the collision constraints are fulfilled. 
With fixed timesteps, the number of waypoints $K$ corresponds to the time duration of the trajectory.
$K$ is known as the \emph{makespan}.
Our planner minimizes $K$ to produce short trajectories.

\subsection{Unlabeled Planner}

We model unlabeled planning as a variant of the \emph{unlabeled Multi-Agent Path-Finding} (MAPF) problem.
We are given an undirected connected graph of the environment ${\mathcal{G}_{E}=(\mathcal{V}_E,\mathcal{E}_E)}$, where each vertex $v\in\mathcal{V}_E$ corresponds to a location in $\cF$ and each edge $(u,v)\in\mathcal{E}_E$ denotes that there is a linear path in $\cF$
connecting $u$ and $v$.
Obstacles are implicitly modeled by not including a vertex in $\mathcal{V}_E$ for each cell that contains an obstacle.
We assume that there exists a vertex $v_s^i\in\mathcal{V}_E$ corresponding to each start location $s^i$ and that there exists a vertex $v_g^i\in\mathcal{V}_E$ for each goal location $g^i$.
At each discrete timestep, a robot can either wait at its current vertex or traverse an edge.
For the following formulation, we assume that the locations corresponding to the vertices are in a grid world and that $z(\cdot)$ and $xy(\cdot)$ map a vector to its $z$ and $x,y$ components, respectively.
Our goal is to find paths $p^i$, such that the following properties hold:
\vspace{0.25cm}
\begin{propertylist}
  \item Each robot starts at its start vertex:
  $\forall i: x^i_0=s^i$.
  
    \item Each robot ends at its goal vertex:
  $\forall i: x^i_K=g^{\phi (i)}$.
  
  \item At each timestep, each robot either stays at its current position or traverses an edge:
  $\forall k, \forall i$: $x^i_{k} = x^i_{k+1}$ or 
  $\exists \ (u,v)\in\mathcal{E}_E$ s.t. $u$ and $v$ 
  correspond to $x^i_k$ and $x^i_{k+1}$.
    
  \item No robots occupy the same location at the same time 
  (\emph{vertex collision}):
  $\forall k, \forall i\neq j$: $x^i_k \neq x^j_k$.
  
  \item No robots traverse the same edge in opposite directions 
  (\emph{edge collision}):
  $\forall k, \forall i\neq j$: $x^i_k \neq x^j_{k+1}$ or $x^j_k \neq x^i_{k+1}$.
  
  \item Robots obey downwash constraints when stationary 
  (\emph{downwash vertex collision}):
  $\forall k, \forall i\neq j$ where $xy(x^i_k) = xy(x^j_k)$: $|z(x^i_k)-z(x^j_k)| \ge 2 r_z$.
  
  \item Robots obey downwash constraints while traversing an edge 
  (\emph{downwash edge collision}):
  $\forall k, \forall i\neq j$ where $xy(x^i_k)=xy(x^j_{k+1}), xy(x^j_k)=xy(x^i_{k+1})$:
  $|z(x^i_k) - z(x^j_{k+1})|\ge 2 r_z$ or $|z(x^j_k)-z(x^i_{k+1})| \ge 2 r_z$.
\end{propertylist}
\vspace{0.25cm}
We consider a solution optimal if the makespan $K$ is minimal.
If only the first five properties are considered and $K$ is given, 
unlabeled MAPF can be solved in polynomial time 
by reduction to a maximum-flow problem in a larger graph, derived from $\mathcal{G}$,
known as a \emph{time-expanded flow-graph}~\cite{DBLP:journals/corr/abs-1204-5717}.
This graph, denoted by $\mathcal{G}_F$,
contains $O(K \cdot |\mathcal{V}_E|)$ vertices and is constructed such that
a flow in $\mathcal{G}_F$ represents a solution to the MAPF instance.
This maximum-flow problem can also be expressed as an 
Integer Linear Program (ILP) where each edge is modeled as binary variable indicating its flow
and the objective is to maximize the flow subject to flow conservation constraints~\cite{DBLP:conf/icra/YuL13}.
An ILP formulation  allows us to add additional constraints for P6 and P7.

We build the time-expanded flow-graph ${\mathcal{G}_F=(\mathcal{V}_F, \mathcal{E}_F)}$ as intermediate step to formulate the ILP.
Compared to the existing detailed discussions~\cite{DBLP:journals/corr/abs-1204-5717,DBLP:conf/icra/YuL13,DBLP:conf/atal/MaK16},
we add additional annotations $con: \mathcal{E}_F \mapsto 2^{\mathcal{E}_F}$ to some of the edges such that $con(e)$ is the set of edges with which $e$ is in conflict under the downwash model.
For each timestep $k$ and vertex ${v\in\mathcal{V}_E}$ we add two vertices $u^v_k$ and $w^v_k$ to $\mathcal{V}_F$ and create an edge connecting them.
\newpage
For each timestep $k$ and edge $(v_1,v_2)\in\mathcal{E}_E$, we create a ``gadget'' connecting $u^{v_1}_k, u^{v_2}_k, w^{v_1}_k$, and $w^{v_2}_k$.
As shown in Fig~\ref{fig:flowgraph:gadget}, the ``gadget'' disallows agents to swap their positions in one timestep, thus enforcing P5.
Furthermore, we connect consecutive timesteps with additional edges $(w^v_k, u^v_{k+1})$ (green edges in Fig.~\ref{fig:flowgraph:example}) to enforce P4.
Additionally we add vertices $source$ and $sink$, which are connected to vertices 
$\{ u^{v_s^i}_0 : \forall i \}$ and $\{ w^{v_g^i}_K : \forall i \}$
respectively.
If a maximum flow is computed on this graph, the flow describes a path for each robot, fulfilling P1--P5.

Consider vertices $v,v'\in\mathcal{V}_E$ that, if simultaneously occupied, would violate P6.
Those vertices map to helper edges $e_k=(w^v_k, u^v_{k+1})$ and ${e_k'=(w^{v'}_k, u^{v'}_{k+1})}$ in $\mathcal{G}_F$ for all~$k$ (green edges in Fig.~\ref{fig:flowgraph:example}).
In that case we insert $e'_k$ into $con(e_k)$ and $e_k$ into $con(e'_k)$ for all~$k$.
Similarly, consider $(v_1,v_2)\in\mathcal{E}_E$ and $(v_1',v_2')\in\mathcal{E}_E$ that violate P7.
These edges map to helper edges $e_k$ and $e_k'$ in $\mathcal{G}_F$ as part of the gadget for all $k$ (blue edges in Fig.~\ref{fig:flowgraph:example}).
As before we insert $e'_k$ into $con(e_k)$ for all $k$ and vice versa.

For each edge $(u,v)\in\mathcal{E}_F$, we introduce a binary variable $z_{(u,v)}$.
The ILP can be formulated as follows:

\begin{equation}\begin{split}
\text{maximize} \; & \sum_{(source,v)\in\mathcal{E}_F} \! \! \! \! z_{(source,v)} \\
\text{subject to} \;
& \sum_{(u,v)\in\mathcal{E}_F} \! \! \! \! z_{(u,v)} = \! \! \! \! \sum_{(v,w)\in\mathcal{E}_F} \! \! \! \! z_{(v,w)}
\ \forall v \in \mathcal{V}'_F \\
& z_e + \sum_{e'\, \in\, con(e)} \! \! \! \! z_{e'} \leq 1
\quad \forall e\in\mathcal{E}_F \\
\end{split}\end{equation}
where $\mathcal{V}'_F = \mathcal{V}_F \setminus \{source, sink\}$.
The first constraint enforces flow conservation, and thus P3--P5. The second constraint enforces P6--P7.
P1 and P2 are implicitly enforced by construction of the flow graph.
A solution to the ILP 
assigns a flow to each edge.
We can then easily create the path $p^i$ for each robot by setting $t_k=k \Delta t$ for any $\Delta t > 0$, and $x^i_k$ based on the flow in $\mathcal{G}_F$.

In order to find an optimal solution for an unknown $K$, we use a two-step approach.
First, we find a lower bound for $K$
by ignoring P6 and P7.
We search the sequence ${K = 1, 2, 4, 8, \dots}$ for a feasible $K$,
and then perform a binary search to find the minimal feasible $K$, which we denote as $LB(K)$.
Because we ignore the downwash constraints, we can check the feasibility in polynomial time using the Edmonds-Karp algorithm on the time-expanded flow-graph.
Second, we execute a linear search 
starting from $LB(K)$,
solving the fully constrained ILP.
In practice, we have found that the lower bound $LB(K)$ is sufficiently close to the final $K$ such that a linear search is faster compared to another modified binary search using the ILP.

\section{CONTINUOUS REFINEMENT STAGE}
In the continuous refinement stage, we convert the waypoint sequences
$p^i$ generated by the discrete planner into smooth trajectories $f^i$.
We use the discrete plan to partition the free space $\cF$ 
such that each robot solves an independent smooth trajectory optimization problem
in a region that is guaranteed to be collision-free.

\subsection{Spatial Partition}
The continuous refinement method begins by finding \emph{safe corridors} within the free space $\cF$ for each robot.
The safe corridor for robot $r^i$ is a sequence of convex polyhedra
$\cP^i_k, k \in \{1 \dots K\}$,
such that, if each $r^i$ travels within $\cP^i_k$ during time interval $[t_{k-1}, t_{k}]$,
both robot-robot and robot-obstacle collision avoidance are guaranteed.
For robot $r^i$ in timestep $k$, the safe polyhedron $\cP^i_k$ is the intersection of:
\begin{itemize}
	\item $N - 1$ half-spaces separating $r^i$ from $r^j$ for $j \neq i$
	\item $N_{obs}$ half-spaces separating $r^i$ from $\cO_1 \dots \cO_{N_{obs}}$.
\end{itemize}
We separate $r^i$ from $r^j$ by finding a separating hyperplane
$(\alpha \ij_k, \beta \ij_k)$ such that:
\begin{equation}\begin{split}
\ell^i_k & \subset \{ x : {\alpha \ij_k}^T x < \beta \ij_k \} \\
\ell^j_k & \subset \{ x : {\alpha \ij_k}^T x > \beta \ij_k \}.
\end{split}\end{equation}
While this hyperplane separates $\ell^i_k$ and $\ell_j^k$,
it does not account for the robot ellipsoids.
Without loss of generality, suppose the hyperplanes are given in the normalized form
where $\|\alpha \ij_k\|_2 = 1$.
Then we accomodate the ellipsoids by shifting each hyperplane according to its normal vector:
\begin{equation}\begin{split}
{\beta \ij_k}' &= \beta \ij_k - \|E \alpha \ij_k\| \\
{\beta \ji_k}' &= \beta \ij_k + \|E \alpha \ij_k\|
\label{offset}
\end{split}\end{equation}
where $E = \diag(r_x, r_y, r_z)$ is the ellipsoid matrix.
Robot-obstacle separating hyperplanes are computed similarly,
except we use a different ellipsoid $E_{obs}$ for obstacles
to model the fact that downwash is only important for robot-robot interactions,
and we shift the hyperplanes such that they touch the obstacles.

In our implementation, we require that the obstacles $\cO_i$ are bounded convex polytopes
described by vertex lists.
Line segments are also convex polytopes described by vertex lists.
Computing a separating hyperplane between two disjoint convex polytopes
${\Psi = \conv (\psi_1 \dots \psi_{m_\Psi})}$
and
${\Omega = \conv (\omega_1 \dots \omega_{m_\Omega})}$,
where $\conv$ denotes the convex hull,
can be posed
as an instance of the hard-margin support vector machine (SVM) problem \cite{svm}.
However, the ellipsoid robot shape alters the problem:
for a separating hyperplane with unit normal vector $\alpha$,
the minimal safe margin is $2 \|E \alpha\|_2$.
Incorporating this constraint in the standard hard-margin SVM formulation yields
a slightly modified version of the typical SVM quadratic program:
\begin{equation}\begin{split}
\text{minimize} \quad & \alpha^T E^2 \alpha \\
\text{subject to} \quad
& \alpha^T \psi_i - \beta \leq 1 \quad \text{for } i \in 1 \dots m_\Psi \\
& \alpha^T \omega_i - \beta \geq 1 \quad \text{for } i \in 1 \dots m_\Omega
\label{ellipsoid-svm}
\end{split}\end{equation}
We solve a problem of this form for each robot-robot and robot-obstacle half-space
to yield the safe polyhedron $\cP^i_k$ in the form of a set of linear inequalities.
Note that the safe polyhedra need not be bounded
and that ${\cP^i_k \cap \cP^i_{k+1} \neq \emptyset}$ in general.
In fact, the overlap between consecutive $\cP^i_k$
allows the smooth trajectories to deviate significantly from the discrete plans,
which is an advantage when the discrete plan is far from optimal.

\subsection{B\'ezier Trajectory basis}
After computing safe corridors, we plan a smooth trajectory $f^i(t)$
for each robot, contained within the robot's safe corridor.
We represent these trajectories as piecewise polynomials with one piece per time interval $[t_k, t_{k+1}]$.
Piecewise polynomials are widely used for trajectory planning:
with an appropriate choice of degree and number of pieces,
they can represent arbitrarily complex trajectories
with an arbitrary number of continuous derivatives.

We denote the $k\th$ piece of robot $i$'s piecewise polynomial trajectory as $f^i_k$.
We wish to constrain $f^i_k$ to lie within the safe polyhedron $\cP^i_k$.
However, when working in the standard monomial basis, i.e.
when the decision variables are the $a_i$ in the expression
$$
p(t) = a_0 + a_1 t + a_2 t^2 + \cdots + a_D t^D,
$$
bounding the polynomial inside a convex polyhedron is not a convex constraint.
Instead, we formulate trajectories as B\'ezier curves.
A \mbox{degree-$D$} B\'ezier curve is defined by a sequence of $D+1$
\emph{control points} $y_i \in \R^3$ and a fixed set of Bernstein polynomials, such that
\begin{equation}\begin{split}
f(t) = b_{0,D}(t) y_0 + b_{1,D}(t) y_1 + \cdots + b_{D,D}(t) y_D
\end{split}\end{equation}
where each $b_{i,D}$ is a degree-$D$ Bernstein polynomial
with coefficients\footnote{
The canonical Bernstein polynomials are defined over the time interval $[0,1]$,
but they are easily modified to span our desired time interval.
}
given in \cite{joybernstein}.
The curve begins at $y_0$ and ends at $y_D$.
In between, it does not pass \emph{through} the intervening control points,
but rather is guaranteed to lie in the convex hull of all control points.
Thus, when using B\'ezier control points as decision variables instead of monomial coefficients,
constraining the control points to lie inside a safe polyhedron
guarantees that the resulting polynomial will lie inside the polyhedron also.
We define $f^i$ as a $K$-piece, degree-$D$ B\'ezier curve 
and denote the $d\th$ control point of $f^i_k$
as $y^i_{k,d}$.
The degree parameter $D$ must be sufficiently high to ensure continuity at the user-defined continuity level $C$.

\subsection{Optimization Problem}
The set of B\'ezier curves that lie within a given safe corridor
describes a family of feasible solutions to a single robot's planning problem.
We select an optimal trajectory by minimizing a weighted combination
of the integrated squared derivatives:
\begin{equation}\begin{split}
\cost(f^i) =
\sum_{c = 1}^C \gamma_c \int_0^T \left\| \frac{d^c}{dt^c} f^i(t) \right\|_2^2 dt
\label{cost}
\end{split}\end{equation}
where the $\gamma_c \geq 0$ are user-chosen weights on the derivatives.
A typical choice in our experiments is to penalize acceleration and snap equally.
As an input to the trajectory optimization stage,
we require the user to supply an initial guess of the duration $\Delta t$ of each timestep,
such that ${T = K \Delta t}$.

Our decision variable $\vy$ consists of all control points 
for $f^i$ concatenated together:
\begin{equation}
\vy = \bmat{
{y^i_{1,0}}^T \dots {y^i_{1,D}}^T, & \dots, & 
{y^i_{K,0}}^T \dots {y^i_{K,D}}^T}^T
\end{equation}
The objective function \eqref{cost} is a quadratic function of $\vy$,
which can be expressed in the form:
\begin{equation}\begin{split}
\cost(f^i) = \vy^T (B^T Q B) \vy
\label{quadcost}
\end{split}\end{equation}
where $B$ is a block-diagonal matrix transforming control points into polynomial coefficients,
and the formula for $Q$ is given in \cite{richter}.
The start and goal position constraints,
as well as the continuity constraints between successive polynomial pieces,
can be expressed as linear equalities.
Thus, we solve the quadratic program:
\begin{equation}\begin{split}
\text{minimize} \quad & \vy^T (B^T Q B) \vy \\
\text{subject to} \quad
& y^i_{k,d} \in \cP^i_k \quad \forall \; i, k, d \\
& f^i(0) = s^i,\ f^i(T) = g^{\phi(i)} \\
& f^i \text{ continuous up to derivative $C$} \\
& \frac{d^c}{dt^c} f^i(t) = 0 \quad \forall \; c > 0, \; t \in \{0, T\}
\label{quadprog}
\end{split}\end{equation}
It is important to note that this quadratic program may not always have a solution
due to our conservative assumptions regarding velocity profiles.
In these cases, we fall back on a solution that follows the discrete plan exactly,
coming to a complete stop at corners.
Details of this solution are given in \cite{tang}.

The corridor-constrained B\'ezier formulation presents one notable shortcoming:
for a given safe polyhedron $\cP^i_k$, there exist degree-$D$ polynomials
that lie inside the polyhedron but cannot be expressed as a B\'ezier curve
with control points that are contained within $\cP^i_k$.
Empirical exploration of B\'ezier curves suggests that this problem is most significant
when the desired trajectory is near the faces of the polyhedron rather than the center.
Further research is needed to characterize this issue more precisely.

\begin{figure}[b]
\centering
\begin{subfigure}[b]{0.35\columnwidth}
\includegraphics[scale=1.8]{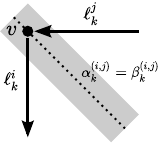}
\caption{}
\end{subfigure}
\hspace{0.5in}
\begin{subfigure}[b]{0.28\columnwidth}
\includegraphics[scale=1.8]{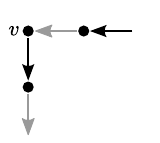}
\caption{}
\end{subfigure}
\caption{Illustration of discrete plan postprocessing.
(a)
In timestep $k$, robot $r^j$ arrives at a graph vertex $v$ and robot $r^i$ leaves $v$.
The separating hyperplane between $\ell^i_k$ and $\ell^j_k$ (with ellipsoid offset shaded in grey)
prevents both robots from planning a trajectory that passes through $v$.
(b) Subdivision of discrete plan ensures that this situation cannot occur.
}
\label{subdivision}
\end{figure}

\begin{table*}[t]
\centering
\caption{Runtime for different examples and safety distances, see Section \ref{sec:exp:runtime}. All times are given in seconds.
}
\begin{tabular}{c|c|c|c|c||c|c|c|c|c||c|c|c|c|c}
\multicolumn{5}{c||}{} & \multicolumn{5}{c||}{Discrete} & \multicolumn{5}{c}{Continuous} \\
Example & $N$ & $N_{obs}$ & Grid Size & $r_z$ & $LB(K)$ & $t_{LB}$ & $t_{ILP}$ & $K$ & $t_{dis}$ & $N_{redObst}$ & $t_1$ & $t_{1(hp)}$ & $t_{1(qp)}$ & $t_{con}$\\
\hline
\hline
\multirow{2}{*}{USC} & \multirow{2}{*}{32} & \multirow{2}{*}{61} & \multirow{2}{*}{$13\times 13\times 5$} & 0.3 & \multirow{2}{*}{15} & \multirow{2}{*}{2.3} & 19   & 17 & 39  & \multirow{2}{*}{9} & \multirow{2}{*}{2.7} & \multirow{2}{*}{0.6} & \multirow{2}{*}{2.1} & \multirow{2}{*}{19}\\
    &    & &                      & 0.9 & & & 17   & 17 & 32 & & & &\\
\hline
\multirow{2}{*}{Maze50} & \multirow{2}{*}{50} & \multirow{2}{*}{441} & \multirow{2}{*}{$20\times 13\times 5$} & 0.3 & \multirow{2}{*}{26} & \multirow{2}{*}{8.6} & 51   & 26 & 60  & \multirow{2}{*}{43} & \multirow{2}{*}{8.6} & \multirow{2}{*}{2.8} & \multirow{2}{*}{5.8} & \multirow{2}{*}{57}\\
    &    & &                      & 0.9 & & & 67   & 26 & 76 & & & &\\
\hline
\multirow{2}{*}{Sort200} & \multirow{2}{*}{200} & \multirow{2}{*}{1320} & \multirow{2}{*}{$29\times 29\times 5$} & 0.3 & \multirow{2}{*}{19} & \multirow{2}{*}{101} & 438 & 19 & 541 & \multirow{2}{*}{94} & \multirow{2}{*}{36} & \multirow{2}{*}{20} & \multirow{2}{*}{16} & \multirow{2}{*}{239}\\
    &    & &                      & 0.9 & & & 615   & 19 & 722 & & & &
\end{tabular}
\label{tab:runtime}
\end{table*}

\subsection{Iterative Refinement}
Solving \eqref{quadprog} for each robot converts the discrete plan
into a set of smooth trajectories that are locally optimal given the spatial decomposition.
However, these trajectories are not globally optimal.
In our experiments, we found that the smooth trajectories sometimes
lie quite far away from the original discrete plan.
Motivated by this observation, we implement an iterative refinement stage
where we use the smooth trajectories to define a new spatial decomposition,
and use the same optimization method to solve for a new set of smooth trajectories.

For time interval $k$, we sample $f^i_k$
at $S$ evenly-spaced points in time to generate a set of points $\cS^i_k$.
The number of sample points $S$ is a user-specified parameter, set to $S = 32$ in our experiments.
We then compute the separating hyperplanes as before, except we separate
$\cS^i_k$ from $\cS^j_k$ instead of $\ell^i_k$ from $\ell^j_k$.
This problem is also a (slightly larger) ellipsoid-weighted support vector machine instance.
While the sample points $\cS^i_k$ are not a complete description of $f^i_k$,
$\cS^i_k$ is guaranteed to be linearly separable from $\cS^j_k$
for $i \neq j$, because the polynomial pieces $f^i_k, f^j_k$ lie inside
their respective disjoint polyhedra $\cP^i_k$, $\cP^j_k$.

These new safe corridors are roughly ``centered'' on the smooth trajectories,
rather than on the discrete plan.
Intuitively, iterative refinement provides a chance for the smooth trajectories to move further
towards a local optimum that was not feasible under the original spatial decomposition.

Iterative refinement can be classified as an anytime algorithm.
If a solution is needed quickly, the original set of $f^i$ can be obtained in a few seconds.
If the budget of computational time is larger, iterative refinement
can be repeated until the quadratic program cost \eqref{quadcost} converges.

The user-supplied timestep duration $\Delta t$ directly affects the magnitudes of dynamic quantities
such as acceleration and snap that are constrained by the robot's actuation limits.
In the case that the final refined trajectories $f^i$ violate some constraint,
we can apply a uniform temporal scaling to all trajectories.
For quadrotors, as the temporal scaling goes to infinity,
the actuator commands are guaranteed to approach a hover state~\cite{mellingersnap},
so kinodynamically feasible trajectories can always be found.

\subsection{Discrete Postprocessing}

Our grid-based MAPF discrete planner produces waypoints $p^i$ that require some postprocessing
to ensure that they satisfy the collision constraints \eqref{eq:collision} under arbitrary velocity profiles.
In particular, we must deal with the case when one robot $r^i$ arrives at a vertex $v \in \mathcal{V}_E$
in the same timestep $k$ when another robot $r^j$ leaves $v$.
This situation creates a conflict where neither robot's smooth trajectory can pass through $v$,
as illustrated in Fig. \ref{subdivision}.
We ensure that this situation cannot happen by dividing each discrete line segment in half.
In the subdivided discrete plan, odd timesteps exit a graph-vertex waypoint and arrive at a segment-midpoint waypoint,
while even timesteps exit a segment-midpoint waypoint and arrive at a graph-vertex waypoint.
Under this subdivision, the conflict cannot occur.

In our experiments, we noticed that the continuous trajectories typically
experience peak acceleration at ${t = 0}$ and ${t = T}$ due to the requirement of accelerating to/from a complete stop.
We add an additional wait state at the beginning and end of the discrete plans to reduce the acceleration peak.

\section{EXPERIMENTS}
We implement the discrete planner in C++ using the Boost Graph library for maximum flow computation and Gurobi 7.0 as the ILP solver.
The continuous refinement stage is implemented in Matlab.
We convert adjacent grid-cell obstacles into $N_{redObst}$ larger boxes using a greedy algorithm.
To compute separating hyperplanes for the safe corridors, 
our method requires solving $O(KN^2 + KN_{redObs}N)$ small ellipsoid-weighted SVM problems.
For these problems, we use the \mbox{CVXGEN} package~\cite{cvxgen} to generate C code optimized for the exact quadratic program specification \eqref{ellipsoid-svm}.
The per-robot trajectory optimization quadratic programs \eqref{quadprog} are solved using Matlab's \texttt{quadprog} solver.
Since these problems are independent, this stage can take advantage of up to $N$ additional processor cores.

In our experiments, we use a polynomial degree ${D = 7}$ and enforce continuity up to the fourth derivative (${C = 4}$).
We evaluate our method in simulation and on the Crazyswarm --- a swarm of nano-quadrotors~\cite{crazyswarm}.

\begin{figure*}
\centering
\begin{subfigure}[t]{0.49\textwidth}
\centering
\includegraphics[width=0.9\textwidth,clip,trim={0, -0.4in, 0, 0}]{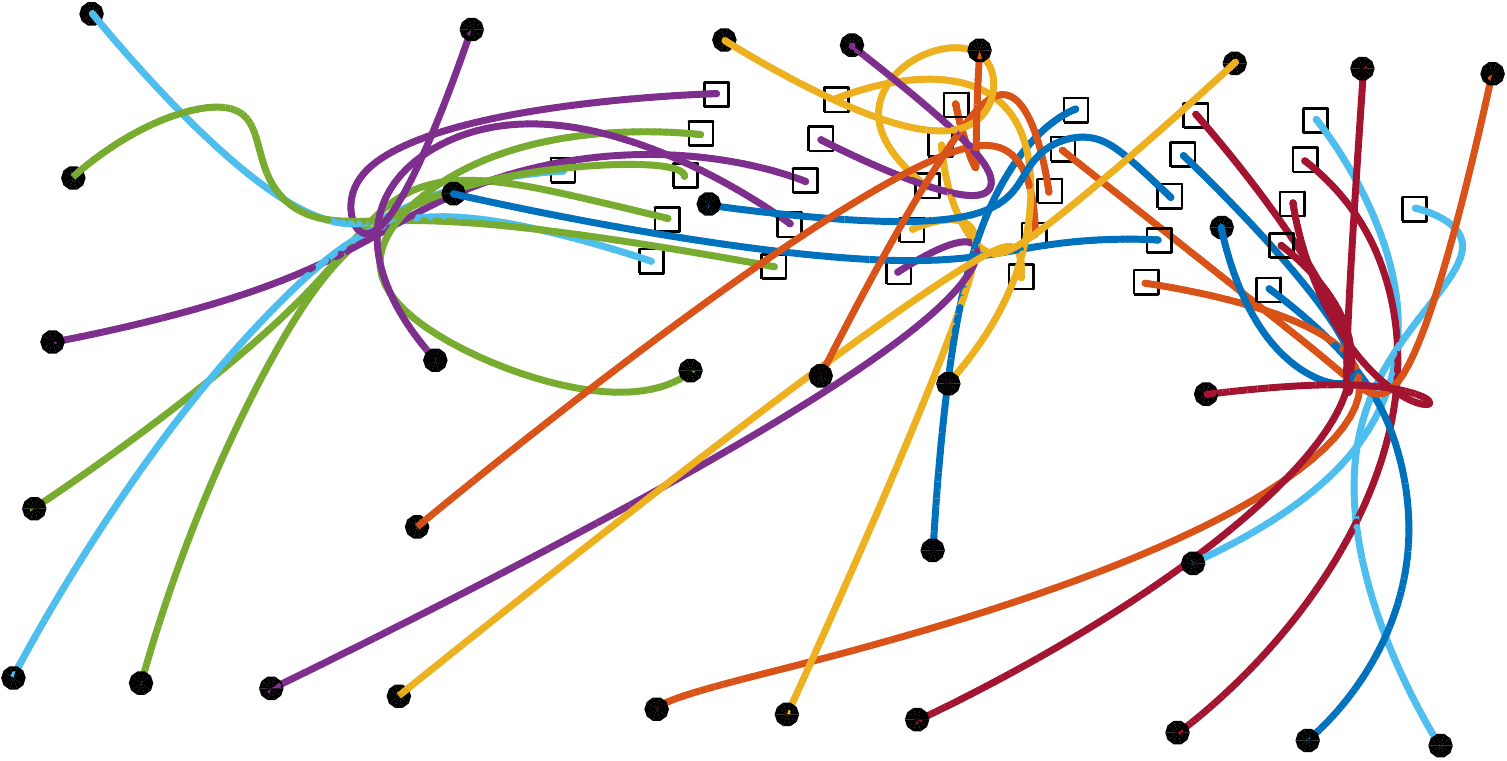}
\caption{Full 32-robot trajectory plan after six iterations of refinement.
The start and end positions are marked by squares and filled circles, respectively.
The obstacles are not shown for clarity.
}
\label{fig:usc:traj}
\end{subfigure} \hfill
\begin{subfigure}[t]{0.49\textwidth}
\centering
\includegraphics[width=1.0\textwidth]{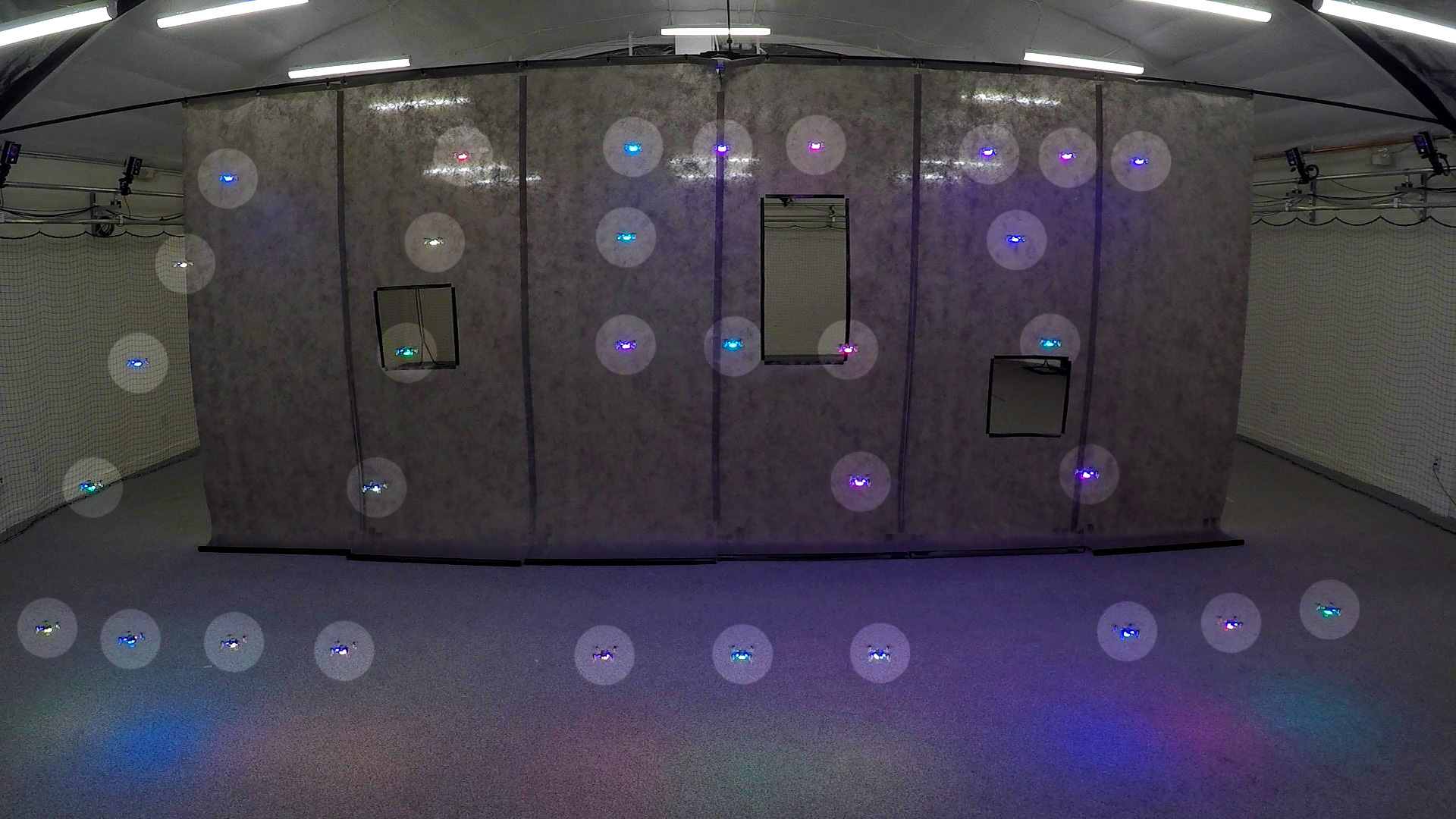}
\caption{Picture of the final configuration after the test flight. A video is available as supplemental material.
}
\label{fig:usc:warehouse}
\end{subfigure}
\caption{Formation change example where quadrotors fly from an $xy-$plane grid formation to a goal configuration spelling ``USC'' while avoiding obstacles.
}
\label{fig:usc}
\end{figure*}

\begin{figure*}
\centering
\begin{subfigure}[b]{0.32\textwidth}
\centering
\includegraphics[width=0.75\textwidth]{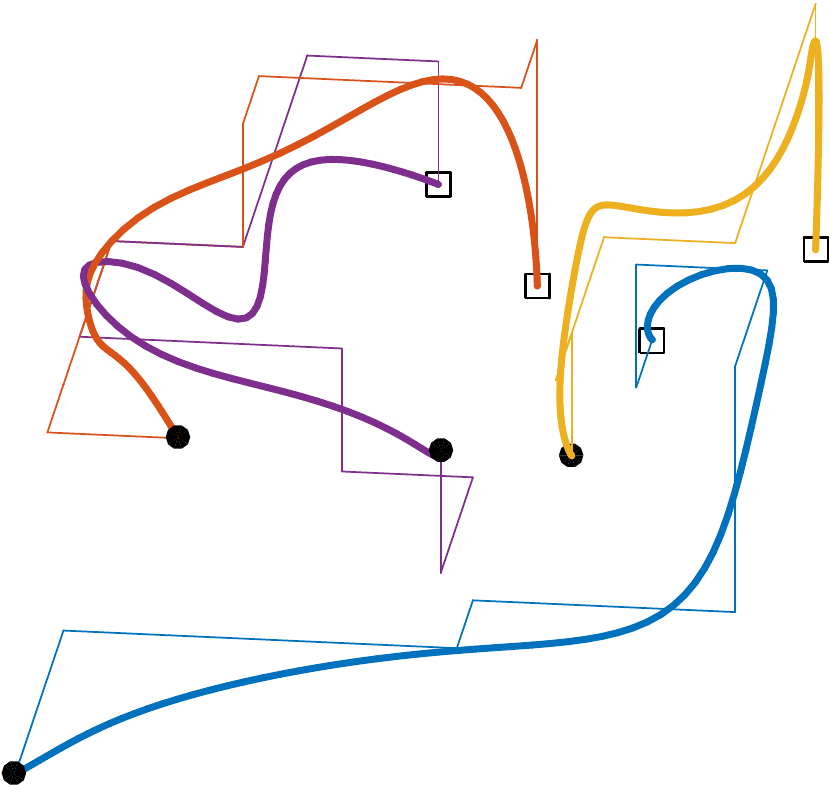}
\caption{1 iteration}
\label{fig:refinement:iter1}
\end{subfigure}
\hfill
\begin{subfigure}[b]{0.32\textwidth}
\centering
\includegraphics[width=0.75\textwidth]{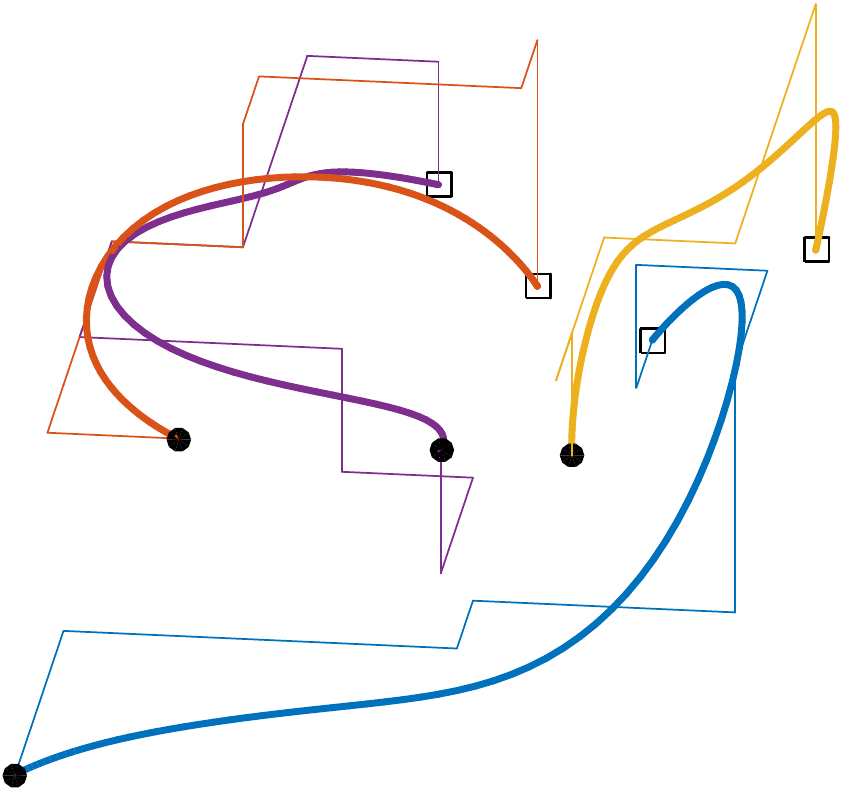}
\caption{2 iterations}
\label{fig:refinement:iter6}
\end{subfigure}
\hfill
\begin{subfigure}[b]{0.32\textwidth}
\centering
\includegraphics[width=0.75\textwidth]{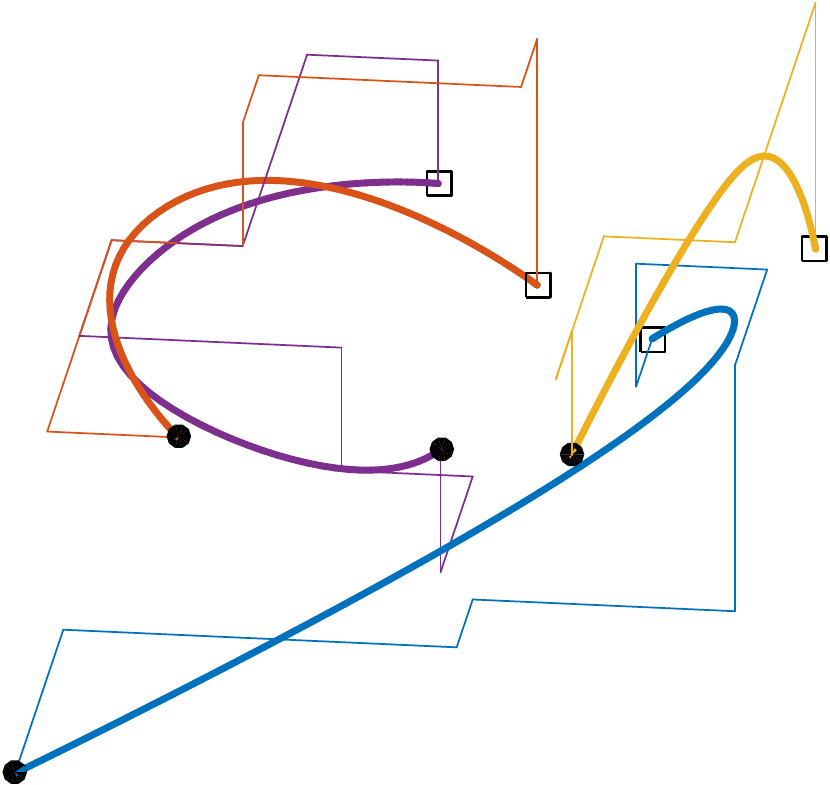}
\caption{6 iterations}
\label{fig:refinement:iter12}
\end{subfigure}
\caption{Subset of results for the example shown in \figref{usc} after different numbers of refinement iterations.
Fine lines represent the discrete plans $p^i$; heavy curves represent the continuous trajectories $f^i$.
The remaining 28 robots are hidden for clarity.
}
\label{fig:refinement}
\end{figure*}

\subsection{Downwash Characterization}
In order to determine the ellpsoid radii $E$, we executed several flight experiments.
For $r_z$, we fly two quadrotors directly on top of each other and record the average position error of both quadrotors at \SI{100}{Hz} for varying distances between the quadrotors.
We noticed that high controller gains lead to very low position errors even in this case, but can cause fatal crashes when the quadrotors are close. We determined $r_z=\SI{0.3}{m}$ to be a safe vertical distance.
For the horizontal direction, we use
${r_x=r_y=\SI{0.12}{m}}$.
We set $E_{obs}$ to a sphere of radius \SI{0.15}{m} based on the size of the Crazyflie quadrotor.

\subsection{Runtime Evaluation}
\label{sec:exp:runtime}
We execute our implementation on a PC running Ubuntu 16.04, with a Xeon E5-2630 \SI{2.2}{GHz} CPU and \SI{32}{GB} RAM.
This CPU has 10 physical cores, which improves the execution runtime for the continuous portion significantly.
We compute plans for three example problems for 32~to~200 robots navigating in obstacle-rich environments.
Table~\ref{tab:runtime} summarizes the problems and breaks down the observed computation time into component parts.
For the discrete step we report the runtime to find $LB(K)$ ($t_{LB}$), the runtime to solve the ILP with known $K$ ($t_{ILP}$), and the total time for the discrete solver to find paths for each robot ($t_{dis}$).
For the continuous step we report $N_{redObst}$, the runtime for the first iteration ($t_1$), and the total time ($t_{con}$). For the first iteration, we report the time for finding the hyperplanes ($t_{1(hp)}$) and the time for solving the quadratic program ($t_{1(qp)}$).

To investigate the effect of the robot ellipsoid size on computation time, we try each example with two ellipsoid heights:
${r_z = \SI{0.3}{m}}$ corresponding to our experimental results,
and ${r_z = \SI{0.9}{m}}$ as an arbitrary larger safety distance.
These necessitate safety margins of one and three empty grid cells, respectively, in the discrete planner.
We notice that the choice of $r_z$ has little impact on the performance
because there is enough slack in the examples to achieve a specific makespan even with higher safety distances.
Furthermore, the estimated lower bound for $K$ is very close to the actual lowest possible $K$ in our examples, and the runtime for the discrete solver is dominated by solving the ILP.

In the continuous portion,
the balance between computing separating hyperplanes
and solving the per-robot quadratic programs depends on the size of the problem.
For all experiments, an initial solution is found in less than one minute.
In these examples, we executed a total of six refinement iterations,
which was enough for the quadratic program cost~\eqref{quadcost} to converge in all of our experiments.

One of the examples (``USC'') is discussed in more detail in the next section. The supplemental material contains animated simulations for all examples.

\subsection{Flight Test}

We discuss the different steps of our approach on a concrete task with 32 quadrotors.
In this task, the quadrotors begin in a grid in the $x-y$ plane, fly through a wall with three holes, and form the letters ``USC'' in the air.

\begin{figure}
\centering
\includegraphics[width=\columnwidth]{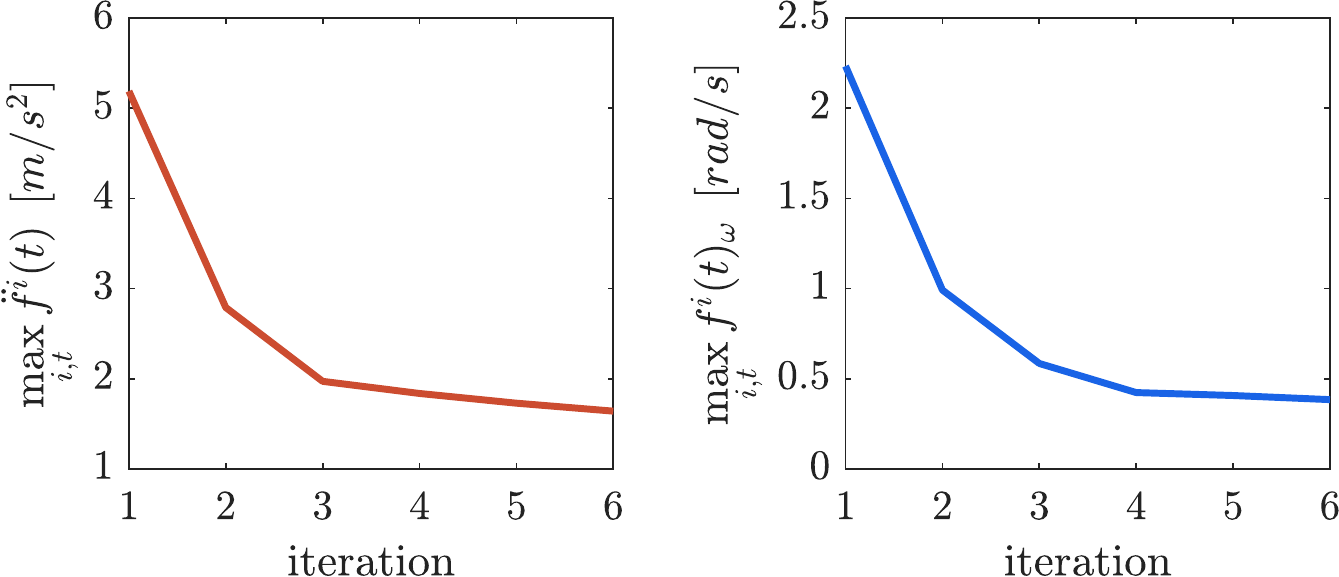}
\caption{
Illustration of worst-case acceleration and angular velocity over all robots
in ``USC'' example
during six iterative refinement cycles.
\emph{Left}: peak acceleration was reduced from \SIrange[range-units=single]{5.2}{1.6}{m/s^2}.
\emph{Right}: peak angular velocity was reduced from \SIrange[range-units=single]{2.2}{0.4}{rad/s}.
}
\label{fig:dynamics_iters}
\end{figure}

The discrete planner plans on a grid of \SI{0.5}{m} side length and finds a solution of $K=17$ timesteps in 40 seconds.
The continuous planner needs three seconds to find the first set of smooth trajectories and finishes six iterations of refinement after 19 seconds.
\figref{dynamics_iters} shows the effect of iterative refinement on the dynamics properties of the trajectories $f^i$.
For each iteration, we take the maximum acceleration and angular velocity
over all robots for the duration of the trajectories.
Iterative refinement results in trajectories with significantly smoother dynamics.
This effect is also qualitatively visible when plotting a subset of the trajectories, as shown in  \figref{refinement}.
The final set of 32 trajectories is shown in \figref{usc:traj}.

We use a swarm of Crazyflie 2.0 nano-quadrotors to execute the trajectories in a space with a physical barrier with windows. The space is \SI{10 x 16 x 2.5}{m} in size and equipped with a VICON motion capture system with 24 cameras.
We upload the planned trajectories to the quadrotors before takeoff, and use the Crazyswarm infrastructure~\cite{crazyswarm} to execute the trajectories.
State estimation and control run onboard the quadrotor, and the motion capture system information is broadcast to the UAVs for localization.
Figure~\ref{fig:usc:warehouse} shows a snapshot of the execution when the quadrotors reached their final state. The executed trajectories can be visualized with long-exposure photography, as shown in \figref{cover}. The supplemental video shows the full trajectory execution.

\section{CONCLUSION}
We presented a trajectory planning method for large quadrotor teams.
Our approach is downwash-aware and thus creates plans where robots can safely fly in close proximity to each other.
We plan trajectories using two independent stages, a discrete stage and a continuous stage.
The presented discrete planner finds a goal assignment for each robot and a path such that the makespan is minimized while avoiding collisions and respecting downwash constraints.
The continuous stage decouples each robot's trajectory planning, allowing easy parallelization 
and improving performance for large teams.
The two-stage architecture supports the use of different discrete multi-agent planners, for example planners where each robot has an assigned goal or task-specific planners.
Iterative refinement offers a user-controlled tradeoff between trajectory quality and computation time.

Our approach can compute safe and arbitrarily smooth trajectories for hundreds of quadrotors in dense environments with obstacles in a few minutes.
The trajectory plan outputs have been tested and executed safely in numerous trials on a team of 32 quadrotors.

In future work, we plan to generalize our method to support arbitrary environments and start and goal locations that are not limited to an underlying grid, by exploring different discrete planning algorithms.
We also plan to investigate performance improvements in both discrete and continuous stages.

\bibliographystyle{IEEEtran}
\bibliography{IEEEabrv,bibliography}{}

\end{document}